\documentclass[runningheads]{llncs}
\usepackage{graphicx}
\usepackage[labelformat=simple]{subcaption}
\usepackage{todonotes}
\graphicspath{ {images/} }
\usepackage{amssymb, amsmath}
\begin{document}
\title{Using Skill Rating as Fitness on the Evolution of GANs}
\author{Victor Costa \and Nuno Louren\c{c}o \and Jo\~{a}o Correia \and Penousal Machado}
\institute{CISUC, Department of Informatics Engineering \\ University of Coimbra, Coimbra, Portugal\\\email{\{vfc, naml, jncor, machado\}@dei.uc.pt}}
\maketitle

\begin{abstract}
Generative Adversarial Networks (GANs) are an adversarial model that achieved impressive results on generative tasks.
In spite of the relevant results, GANs present some challenges regarding stability, making the training usually a hit-and-miss process.
To overcome these challenges, several improvements were proposed to better handle the internal characteristics of the model, such as alternative loss functions or architectural changes on the neural networks used by the generator and the discriminator.
Recent works proposed the use of evolutionary algorithms on GAN training, aiming to solve these challenges and to provide an automatic way to find good models.
In this context, COEGAN proposes the use of coevolution and neuroevolution to orchestrate the training of GANs.
However, previous experiments detected that some of the fitness functions used to guide the evolution are not ideal.

In this work we propose the evaluation of a game-based fitness function to be used within the COEGAN method.
Skill rating is a metric to quantify the skill of players in a game and has already been used to evaluate GANs.
We extend this idea using the skill rating in an evolutionary algorithm to train GANs.
The results show that skill rating can be used as fitness to guide the evolution in COEGAN without the dependence of an external evaluator.

\end{abstract}

\section{Introduction}
Generative models have gained a lot of interest in the past years.
The recent advances in the field contributed with impressive results, mainly in the context of images.
Generative Adversarial Networks (GANs)~\cite{NIPS2014_5423} presented a relevant advance in this context, producing realistic results in several domains.
In the original GAN model, two neural networks, a generator and a discriminator, are competing in a unified training process.
The generator fabricates samples and the discriminator detects if these samples are fake or from an input distribution.

Despite the high-quality results, GANs are hard to train and a trial-and-error strategy is frequently followed to get the expected results.
The challenges with GAN training are commonly related to the balance between the discriminator and the generator.
In this context, the vanishing gradient and the mode collapse are two common problems affecting GANs.
The vanishing gradient leads to stagnation of the training, caused by an imbalance between the forces of the generator and the discriminator.
The mode collapse problem is characterized by the lack of representation of the target distribution used in training.

In order to solve these issues and to achieve better results, different strategies were proposed.
A relevant effort was spent on the design of alternative loss functions to use in the GAN training, originating the proposal of alternative models such as WGAN~\cite{arjovsky2017wasserstein}, LSGAN~\cite{mao2017least}, and RGAN~\cite{jolicoeur-martineau2018}.
Other proposals target the improvement of the architecture used in GANs, defining new modules like in SAGAN~\cite{zhang2018self} or a set of recommendations as in DCGAN~\cite{radford2015unsupervised}.
However, problems like the mode collapse and the vanishing gradient are still present in the training.

The use of evolutionary algorithms to train GANs was recently proposed by some researchers~\cite{al2018towards,costa2019evaluating,costa2019coevolution,garciarena2018evolved,toutouh2019spatial,wang2018evolutionary}.
Techniques such as neuroevolution, coevolution, and Pareto set approximations were used in their models.
The application of evolutionary algorithms in GANs takes advantage of the evolutionary pressure to guide individuals toward convergence, often discarding problematic individuals.

Coevolutionary GAN (COEGAN) proposes the use of neuroevolution and coevolution to orchestrate the training of GANs.
Despite the advances in the training stability, there is still room for improvement in the model.
The experimental evaluation identified that the fitness function can be enhanced to better guide the evolution of the components, mainly regarding the discriminator.
Currently, the discriminator uses the loss function of the respective GAN component.
However, this function displayed a high variability behavior, disrupting the evolution of the population.
The generator uses the Fr\'{e}chet Inception Distance (FID) score, which introduces an external evaluator represented by a trained Inception Network~\cite{szegedy2015going,szegedy2016rethinking}.
Although the good results introduced by the FID score as fitness, the drawbacks are the execution cost and the dependence of an external evaluator.

The FID score is currently the most used metric, but several other metrics were proposed to evaluate the performance of GANs~\cite{borji2019pros,xu2018empirical}.
Metrics such as skill rating was successfully used to evaluate GANs in some contexts~\cite{olsson2018skill}.
Skill rating uses a game rating system to assess the skill of generators and discriminators.
Each generator and discriminator is considered as a player in a game and the pairing between them is designed as a match.
The outcome of the matches is used as input to calculate the skill of each player.

We took inspiration from the use of skill rating to quantify the performance of generators and discriminators in GANs to design a fitness function to be used within COEGAN.
Therefore, we replace the regular fitness used in COEGAN with the skill rating, i.e., the discriminator and the generator use the skill rating metric instead of the loss function and the FID score.
We present an experimental study on the use of this metric, comparing the results with the previous approach used in COEGAN, a random search approach, and with a non-evolutionary model based on DCGAN.
The results evidenced that skill rating provides useful information to guide the evolution of GANs when used in combination with the COEGAN model.
The skill rating is more efficient with respect to execution time and does not compromise the quality of the final results.

The remainder of this paper is organized as follows:
Section \ref{sec:background} introduces the concepts of GANs and evolutionary algorithms, presenting state-of-the-art works using these concepts;
Section \ref{sec:model} presents COEGAN and our approach to use skill rating as fitness;
Section \ref{sec:experiments} displays the experimental results using this approach;
finally, Section \ref{sec:conclusions} presents our conclusions and future work.

\section{Background and Related Works}
\label{sec:background}

Generative Adversarial Networks (GANs)~\cite{NIPS2014_5423} are an adversarial model that have became relevant for presenting high-quality results in generative tasks, mainly on the image domain.
In summary, a GAN is composed of a generator and a discriminator, trained as adversaries by a unified algorithm.
Each component is represented by a neural network and has a role guided by its specific loss function.
The generator has to produce synthetic samples that should be classified as real by the discriminator.
The discriminator should distinguish between fake samples and samples originated from an input distribution.
For this, the discriminator receives a real input distribution for training, such as an image dataset.
The generator is fed with a latent distribution, usually with a lower dimension than the real input distribution, and never directly looks into the real distribution.

In the original GAN model, the loss function of the discriminator is defined as follows:
\begin{equation}
J^{(D)}(D,G) = -\mathbb{E}_{x \sim p_{data}}[\log D(x)] - \mathbb{E}_{z \sim p_z}[\log(1 - D(G(z)))].
\label{eq:discriminator}
\end{equation}

For the generator, the non-saturating version of the loss function is defined by:
\begin{equation}
J^{(G)}(G) = - \mathbb{E}_{z \sim p_z}[\log(D(G(z)))].
\label{eq:generator}
\end{equation}

In Eq. \ref{eq:discriminator}, $p_{data}$ is the real data used as input to the discriminator.
In Eq. \ref{eq:discriminator} and Eq. \ref{eq:generator}, $z$ is the latent space used to feed the generator, $p_z$ is the latent distribution, $G$ is the generator, and $D$ represents the discriminator.

Despite the quality of the results, GANs are hard to train and the presence of stability issues on the training process is frequent.
The vanishing gradient and the mode collapse are two of the most common problems that affect the training of GANs.
The vanishing gradient issue is characterized by a disequilibrium between the forces of the GAN components.
For example, the discriminator becomes too powerful and does not make mistakes when detecting fake samples produced by the generator.
In this case, the progress on the training stagnates.
The mode collapse problem occurs when the generator only partially captures the input distribution used on the discriminator training.
This issue affects the variability and the quality of the created samples.

Several approaches were used to minimize these issues and leverage the quality of the results.
In this context, alternative loss functions were proposed to replace the functions used in the classical GAN model, such as WGAN~\cite{arjovsky2017wasserstein}, LSGAN~\cite{mao2017least}, and RGAN~\cite{jolicoeur-martineau2018}.
Another strategy is to propose architectural changes to the GAN model.
DCGAN~\cite{radford2015unsupervised} proposed a reference architecture for the discriminator and the generator in GANs, describing a set of constraints and rules to achieve better results.
On the other hand, a predefined strategy to progressively grow a GAN during the training procedure was proposed in~\cite{karras2018progressive}.
SAGAN~\cite{zhang2018self} proposed the use of a self-attention module in order to capture the relationship between spatial regions of the input sample.
Although these approaches tried to minimize the problems and produce better results, issues still affect the training of GANs~\cite{arjovsky2017wasserstein,gulrajani2017improved,salimans2016improved}.
Besides, the discovery of efficient models and hyperparameters for the models is not a trivial task, requiring recurrent empirical validation.

Recently, research was conducted to propose the use of evolutionary algorithms to train and evolve GANs~\cite{al2018towards,costa2019evaluating,costa2019coevolution,garciarena2018evolved,toutouh2019spatial,wang2018evolutionary}.
Evolutionary algorithms take inspiration on the mechanism found in nature to evolve a population of potential solutions on the production of better outcomes for a given problem~\cite{sims1994evolving}.
E-GAN~\cite{wang2018evolutionary} uses an evolutionary algorithm to combine three different types of loss functions in the training.
An approach based on the Pareto set approximations was used in \cite{garciarena2018evolved} to model the GAN problem.
Lipizzaner~\cite{al2018towards} proposes the use of spatial coevolution to match generators and discriminators in the training process.
Mustangs~\cite{toutouh2019spatial} unifies the concepts of E-GAN and Lipizzaner in a single model, using different loss functions and spatial coevolution in the solution.

COEGAN uses neuroevolution and coevolution on the training and evolution of GANs.
Despite the advances identified by the experiments, the results also showed that the fitness functions used in the model can be improved.
COEGAN uses the loss function (Eq. \ref{eq:discriminator}) as the fitness for discriminators and the FID score for generators.
The use of better fitness can be helpful for the creation of better models and also avoid the common stability issues when training GANs.
Furthermore, as specified in the FID score, COEGAN uses an external evaluator to quantify the fitness for generators.

Several strategies were proposed to quantify the performance of GANs~\cite{borji2019pros,xu2018empirical}.
Although the FID score is the most used metric to evaluate and compare GANs, alternative approaches can be successfully applied, such as skill rating~\cite{olsson2018skill}.
The skill rating metric for GANs uses the Glicko-2~\cite{glickman2012example} rating system to calculate the performance.
Glicko-2 was also used as comparison criteria between different evolutionary algorithms~\cite{vevcek2014comparison,vevcek2014chess}.

\section{Our Approach}
\label{sec:model}

We present in this section our approach to applying skill rating as fitness in an evolutionary algorithm.
For this, we make use of the previously introduced method called COEGAN~\cite{costa2019evaluating,costa2019coevolution}, adapting the model for our proposal in this paper.
Thus, we firstly introduce in this section the COEGAN algorithm.
After that, we describe the skill rating method and its application in COEGAN.

\subsection{COEGAN}

COEGAN~\cite{costa2019evaluating,costa2019coevolution} proposes the use of neuroevolution and coevolution to train and evolve GANs.
The motivations of COEGAN are to solve the stability issues frequently found when training GANs and also to automatically discover efficient models for different applications.

COEGAN is inspired by DeepNEAT~\cite{miikkulainen2017evolving} to design the model, also using coevolution techniques presented in NEAT applied to competitive coevolution~\cite{stanley2004competitive}.
The genome of COEGAN is represented by a sequential array of genes.
This genome is transformed into a neural network, where each gene directly represents a layer in this network.
The evolution occurs on the architecture and the internal parameters of each layer.
Therefore, the mutation operators were used to add a layer, remove an existing layer, and mutate the internal parameters of a layer.
For the sake of simplicity, in this work we only use convolutional layers in the addition operator.
As in the original COEGAN proposal, crossover was not used in the final model because it introduced instability in the system.

Two separated populations are used in COEGAN: a population of discriminators and a population of generators.
Thus, competitive coevolution was used to design the environment.
In the evaluation phase, individuals are matched following an \textit{all vs. all} strategy, i.e., each generator $G_i$ will be matched against each discriminator $D_j$.
Other strategies can be used, such as \textit{all vs. best}.
However, the \textit{all vs. all} approach achieved the best results, despite the high execution cost with the application.

The selection phase uses a strategy based on NEAT~\cite{neat}.
Therefore, a speciation mechanism is used to promote innovation when evolving the populations.
Fitness sharing adjusts the fitness of the individuals, making the selection proportional to the average fitness of each species.
The species are grouped following the similarity on the genome of the individuals.

The fitness for the discriminator is the respective loss function of the classical GAN model, given by Eq.~\ref{eq:discriminator}.
The fitness of the generator is represented by the Fr\'{e}chet Inception Distance (FID)~\cite{heusel2017gans}, given by:
\begin{equation}
FID(x,g) = ||\mu_x - \mu_g||_2^2 + Tr(\varSigma_x + \varSigma_g - 2(\varSigma_x\varSigma_g)^{1/2}).
\label{eq:fid}
\end{equation}
where $\mu_x$, $\varSigma_x$, $\mu_g$, and $\varSigma_g$ represent the mean and covariance estimated for the real dataset $x$ and fake samples $g$, respectively.
The FID score uses the Inception Network~\cite{szegedy2015going,szegedy2016rethinking}, usually trained with the ImageNet dataset~\cite{russakovsky2015imagenet}, to transform images into a feature space, which is interpreted as a continuous multivariate Gaussian.
The mean and covariance of the two resulting Gaussians for the transformation of real and fake images are applied in Eq.~\ref{eq:fid},

\subsection{Skill Rating}
\label{sec:skill_rating}

In games like chess, it is common to use a rating system to quantify the skill of players.
In this context, the Glicko-2~\cite{glickman2012example} rating system can be used to measure the performance of players given a history of matches.
The Glicko-2 system associates to each player three variables: the rating $r$, the deviation $RD$, and the volatility $\sigma$.
The rating $r$ indicates the actual skill of player after a sequence of matches with other players in a game.
The volatility $\sigma$ represents the expected variability on the rating of a player.
The deviation $RD$ represents the confidence in the player's rating.
A system constant $\tau$ is also used to control the rate of change on the volatility $\sigma$.
Different from $r$, $RD$, and $\sigma$, this parameter is associated with the whole rating system.

All players are initialized with the recommended values of $1500$ for the rating $r$, $350$ for the deviation $RD$ and $0.06$ for the volatility $\sigma$.
These values can be tuned according to the characteristics of the application.
At a fixed time period, the results of all matches between players are stored and used to update the rating $r$, deviation $RD$, and volatility $\sigma$.
It is recommended to use a time span large enough to contain at least $10$ to $15$ games for each player.

The Glicko-2 rating system was previously used on the comparison of evolutionary algorithms~\cite{vevcek2014comparison,vevcek2014chess}.
In this case, different algorithms are executed on a given problem and the solutions found by them are matched to produce the outcome used as input to the Glicko-2 system.
Thus, the algorithms are ranked according to the rating score.

Another application of the Glicko-2 system was to evaluate the performance of GANs~\cite{olsson2018skill}.
In this case, the rating was applied between discriminators and generators of different epochs to calculate the progressive skills of them.
The authors found that skill rating provides a useful metric to relatively compare GANs.

We took inspiration on these use cases of Glicko-2 to apply the system in COEGAN.
The fitness function for discriminators and generators in the COEGAN algorithm was changed to use the skill rating metric computed using Glicko-2.
Therefore, each generator $G_i$ and discriminator $D_j$ have an associated skill rating, represented by $r$, $RD$, and $\sigma$.

At the evaluation phase of the evolutionary algorithm, discriminators and generators are matched to be trained with the GAN algorithm and also to be evaluated for selection and reproduction.
We modeled each evaluation step between a generator and a discriminator as a game to be quantified and applied to the skill rating calculation, composing a tournament of generators against discriminators.
Therefore, as we use the \textit{all vs. all} pairing strategy, each outcome of the match between $(G_i, D_j)$ is stored and used to update the skill rating at the end of each generation.
Inspired by the approach in \cite{olsson2018skill}, we use the following equations to calculate the outcome of a match for the discriminator:
\begin{equation} \label{eq:skill_discriminator_real}
D_j^{real} = \sum_{x \sim p_{data}}th\Big(D_j(x) > 0.5\Big)
\end{equation}
\begin{equation} \label{eq:skill_discriminator_fake}
D_{ij}^{fake} = \sum_{z \sim p_z}th\Big(D_j(G_i(z)) < 0.5\Big)
\end{equation}
\begin{equation} \label{eq:skill_discriminator}
D_{ij}^{wr} = \frac{D_j^{real} + D_{ij}^{fake}}{m + n}
\end{equation}
where $D_j^{real}$ is the win rate of the discriminator with respect to the real data, $D_{ij}^{fake}$ is the rate related to the fake data, $D_{ij}^{WR}$ is the overall win rate of the discriminator $D_j$, $th$ is a threshold function that outputs $1$ when the threshold is met and $0$ otherwise, $D_j$ outputs the probability of the sample to be real, $G_i$ is the generator, $p_{data}$ is the input distribution, $x$ is a sample drawn from the input distribution, $p_z$ is the latent distribution, $z$ is the latent input for the generator, $m$ is the number of real samples, and $n$ is the number of fake samples.
In summary, the win rate for the discriminator is based on the number of mistakes made by it with the real input batch (Eq. \ref{eq:skill_discriminator_real}) and fake data produced by the generator (Eq. \ref{eq:skill_discriminator_fake}).

For the generator, the result is calculated as:
\begin{equation} \label{eq:skill_generator}
G_{ij}^{wr} = 1 - D_{ij}^{wr}
\end{equation}
where $D_{ij}^{wr}$ is the discriminator win rate given by Eq.~\ref{eq:skill_discriminator}.

The win rates of each generator and discriminator are used as input to update the skill rate of the individuals.
Each individual $G_i$ and $D_j$ has a set of outcomes $T^{wr}$, containing the win rate of each match and the skill of the adversarial.
Thus, a generator $G_i$ has a set $T^{wr}_{G_i}$ containing each pair $(G_{ij}^{wr}, D_{j}^{sk})$ for a generation.
A discriminator $D_j$ has a set $T^{wr}_{D_j}$ containing each pair $(D_{ij}^{wr}, G_{i}^{sk})$.
The sets $T^{wr}_{G_i}$ and $T^{wr}_{D_j}$ are used to calculate the new skill rating at the end of the generation, represented by $G_{i}^{sk}$ and $D_{j}^{sk}$, respectively.
It is important to note that the update of the skill rating of a player depends on the skill of the adversary, i.e., win a game from a strong player is more rewarding than to win from a weak player.

We propose in this work the use of skill rating as fitness in COEGAN, represented by the use of $D_{j}^{sk}$ instead of Eq.~\ref{eq:discriminator} for discriminators and $G_{i}^{sk}$ instead of Eq.~\ref{eq:fid} for generators.
Therefore, the fitness functions for discriminators and generators are defined as:
\begin{equation} \label{eq:fitness}
F_{D_j} = r_{D_{j}^{sk}}, \qquad
F_{G_i} = r_{G_{i}^{sk}},
\end{equation}
where $r_{D_{j}^{sk}}$ and $r_{G_{i}^{sk}}$ are the rating $r$ for discriminators and generators, respectively.
At each generation, individuals update the skill rating following these rules.
In the breeding process, the offspring carry the skill rating of their parent.
In this way, we keep track of the progress of individuals through generations, even when mutations occur to change their genome.

Besides the matches between each pair $(G_i, D_j)$, individuals in the current generation can also be matched against individuals from previous generations.
The algorithm can keep track of the best individuals from the last generations to match them against the current individuals in order to ensure the progression of them.
This is also a strategy to avoid the intransitivity problem that occurs in competitive coevolution algorithms.
The intransitivity problem means that a solution $a$ is better than other solution $b$ and $b$ is better than $c$, but it is not guaranteed that $a$ is better than $c$, leading to cycling between solutions during the evolutionary process and harming the progress toward optimal outcomes~\cite{antonio2018coevolutionary,mitchell2006coevolutionary}.
However, this work does not use previous generations in the skill rating calculation.
We leave the evaluation of this strategy for future work.

\section{Experiments}
\label{sec:experiments}

To evaluate the use of skill rating with COEGAN, we conducted an experimental study using the Street View House Numbers (SVHN) dataset~\cite{netzer2011reading}.
The SVHN dataset is composed of digits from $0$ to $9$ extracted from real house numbers.
Therefore, it is a dataset with a structure similar to the MNIST dataset~\cite{lecun1998mnist} used in previous COEGAN experiments, but with more complexity introduced by the use of real images, presenting digits with a variety of backgrounds.
The experiments compare the results of the original COEGAN approach (with the FID score and the loss function as fitness for generators and discriminators), COEGAN with skill rating applied as fitness, a random search approach, and a DCGAN-based architecture.
We also present a comparison between the FID score and the skill rating metric in experiments with the MNIST dataset.

\subsection{Experimental Setup}

\begin{table}[ht]
	\caption{Experimental parameters.}
	\begin{center}\begin{tabular}{c|c}
			\textbf{Evolutionary Parameters} & \textbf{Value} \\
			\hline
			Number of generations & 50 \\
			Population size (generators and discriminators) & 10 \\
			Add Layer rate & 20\% \\
			Remove Layer rate & 10\% \\
			Change Layer rate & 10\% \\
			Output channels range & [32, 256] \\
			Tournament $k_t$ & 2 \\
			FID samples & 2048 \\
			Genome Limit & 4 \\
			Species & 3 \\
			\textbf{Skill Rating Parameters} & \textbf{Value} \\
			\hline
			$r$, $RD$, $\sigma$  & 1500, 350, 0.06 \\
			constant $\tau$ & 1.0 \\
			\textbf{GAN Parameters} & \textbf{Value} \\
			\hline
			Batch size & 64 \\
			Batches per generation & 20 \\
			Optimizer & Adam \\
			Learning rate & 0.001 \\
			Betas & 0.5, 0.999
	\end{tabular}\end{center}
	\label{table:setup}
\end{table}

Table \ref{table:setup} lists the parameters used in our experiments.
These parameters were chosen based on preliminary experiments and the results presented in our previous works~\cite{costa2019evaluating,costa2019coevolution}.
All experiments are executed for $50$ generations.
The number of individuals in the populations of generators and discriminators is $10$.
This number of individuals is enough to achieve the recommended matches to feed the Glicko-2 rating system.
For the variation operators, we use the rates $20\%$, $10\%$, and $10\%$ for the add layer rate, remove layer rate, and change layer rate, respectively.
The number of output channels is sampled using the interval $[32, 256]$.
A tournament with $k_t = 2$ is applied inside each species to select the individuals for reproduction and the algorithm self-adjust to contains $3$ species for the population of generators and discriminators.
The number of samples used to calculate the FID score is $2048$.
To make the experiments comparable, each individual has a genome limited to $4$ genes, the same number of layers used in the DCGAN-based experiments.
Besides, as the DCGAN-based model does not use an evolutionary algorithm, these evolutionary parameters described above are not applied to it.

The initial skill rating parameters used in the experiments are the same suggested by the Glicko-2 system~\cite{glickman2012example}, i.e., the rating $r$, deviation $RD$, and the volatility $\sigma$ are initialized with $1500$, $150$, and $0.06$, respectively.
The system constant $\tau$ was set to $1.0$.
We conduct previous experiments to choose the best $\tau$ for our context.
We found no relevant changes with respect to this parameter.
Nevertheless, experiments focused on the tuning of $\tau$ should be executed to evaluate its effect on our proposal.

All experiments used the original GAN model, i.e., the neural networks are trained with the classical loss functions defined by Eq.~\ref{eq:discriminator} and Eq.~\ref{eq:generator}.
The GAN parameters were chosen based on preliminary experiments and the setup commonly used on the evaluation of GANs~\cite{gulrajani2017improved,karras2018progressive,radford2015unsupervised}.
The batch size used in the training is $64$.
The Adam optimizer~\cite{kingma2015adam} is used with the learning rate of $0.001$, beta 1 of $0.5$, and beta 2 of $0.999$.
Each pairing between generators and discriminators is trained by $20$ batches per generation. As the \textit{all vs. all} is used, each generator and discriminator will be trained for a total of $200$ batches.
For the DCGAN-based experiments, we have a single generator and discriminator.
Therefore, we train them for $200$ batches to keep the results comparable with the COEGAN experiments.

The results are evaluated using the FID score and the skill rating.
For the SVHN dataset, the FID score is based on the Inception Network trained with the SVHN dataset instead of the ImageNet dataset, the same strategy used in the experiments of~\cite{olsson2018skill}.
For the MNIST results, we use the Inception Network trained with the ImageNet dataset.
All results presented in this work are obtained by the average of five executions, with a confidence interval of $95\%$.

\subsection{Results}

\begin{figure}[ht]
	\centering
	\includegraphics[width=0.5\textwidth]{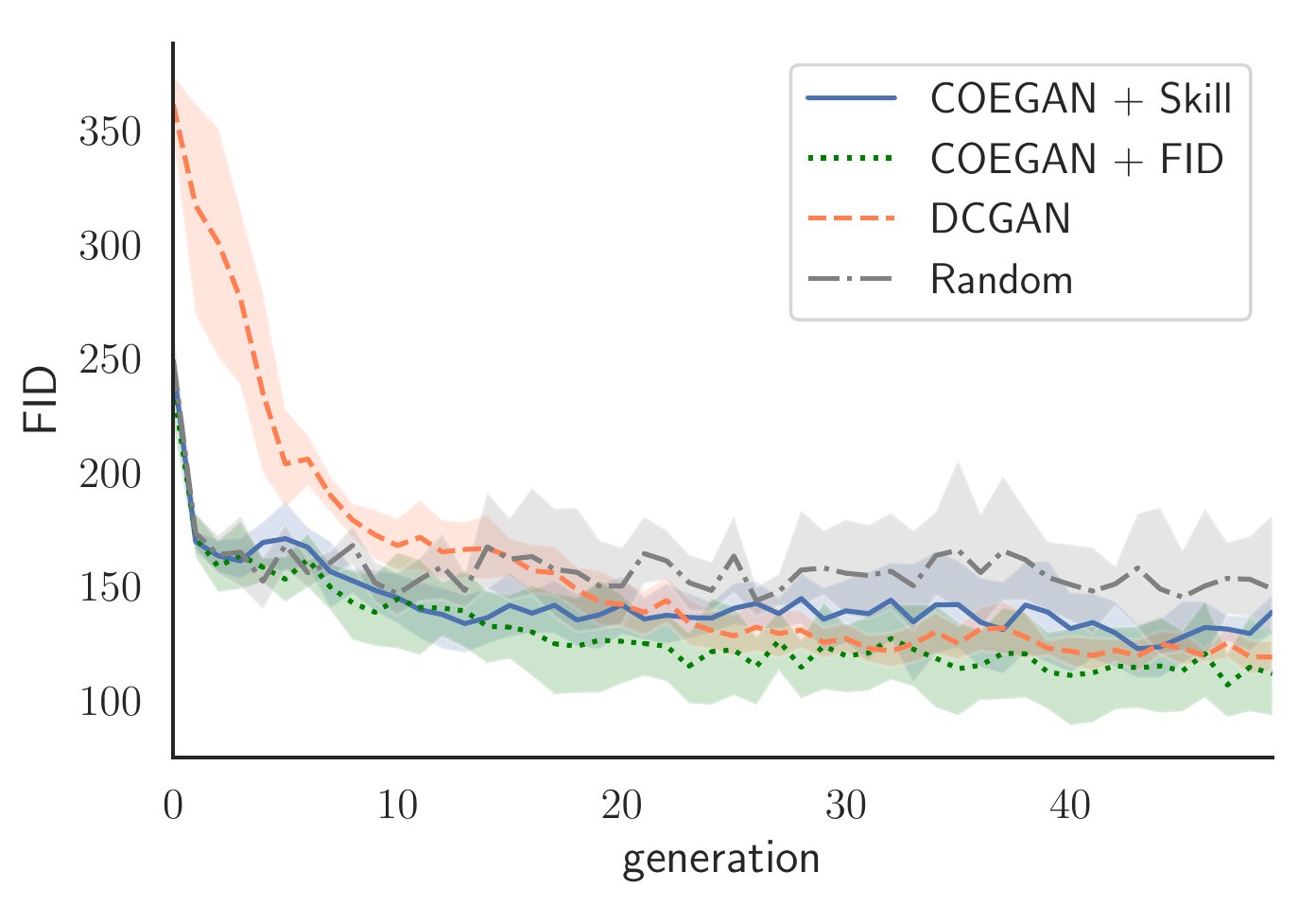}
	\caption{Best FID score for generators with a 95\% confidence interval}
	\label{fig:fid_score}
\end{figure}

Figure~\ref{fig:fid_score} presents the results of the best FID score per generation for the experiments with the SVHN dataset.
We can see that the results for the original COEGAN proposal, i.e., COEGAN guided by the FID and the loss as fitness functions, are still better than the results for COEGAN with the skill rating metric.
However, COEGAN guided by skill rating presented better FID scores than the random search approach.
Thus, this evidences that skill rating provides useful information to the system, presenting evolutionary pressure to the individuals in the search of efficient models.
Moreover, COEGAN with the FID score as fitness outperforms the DCGAN-based approach, illustrating the advantages of COEGAN.

We found in the experiments that skill rating sometimes overestimates the score for bad individuals, affecting the final results of the training.
A dataset with the complexity of SVHN may require more training epochs to achieve better outcomes, and the variability introduced by the \textit{all vs. all} pairing may be too much for complex datasets.
Therefore, another approach such as spatial coevolution used in~\cite{al2018towards,toutouh2019spatial} will be considered in further experiments.
Furthermore, the calculation of the match outcome, given by Eq.~\ref{eq:skill_discriminator_real}-\ref{eq:skill_generator}, can be improved to overcome this problem.

\begin{table}
	\caption{FID score of the algorithms used in the experiments with SVHN.}
	\begin{center}\begin{tabular}{c|c}
			\textbf{Algorithm} & \textbf{FID Score} \\
			\hline
			COEGAN + Skill & $135.1\pm9.8$ \\
			COEGAN + FID & $111.7\pm22.1$ \\
			DCGAN-based & $119.0\pm10.1$ \\
			Random search & $148.9\pm30.7$ \\
	\end{tabular}\end{center}
	\label{table:fid}
\end{table}

Table~\ref{table:fid} shows the average FID of the best scores at the last generation for each experiment with the SVHN dataset.
We can see the difference between the FID of the solutions experimented in this work.
As expected, the results for the random search approach is unstable and worse than the others, presenting a high standard deviation.
However, the difference is not big due to the limitations we impose on the experimental parameters.
Experiments adding the possibility of larger networks for COEGAN should be performed to assess the capacity to outperform both the random search and DCGAN approaches by a larger margin.

Despite the inferior results when compared to COEGAN with FID as fitness, the advantage with the skill rating is that we can avoid the use of an external evaluator as in the FID calculation, represented by the Inception Network.
The execution cost of the skill rating metric is also lower than the FID score.
The FID score requires a high number of samples to have a good representation of the data.
In our experiments, we use $2048$ against $64$ on the skill rating calculation ($64$ represents the batch size used in Eq.~\ref{eq:skill_discriminator}).
Furthermore, the Inception Network has a complex architecture and the FID score uses slow procedures in the calculation.
Skill rating uses the own neural network of individuals in the experiments, and the Glicko-2 system is fast to execute.

\begin{figure}[ht]
	\centering
	\begin{subfigure}[t]{.45\textwidth}
		\centering
		\includegraphics[width=\textwidth]{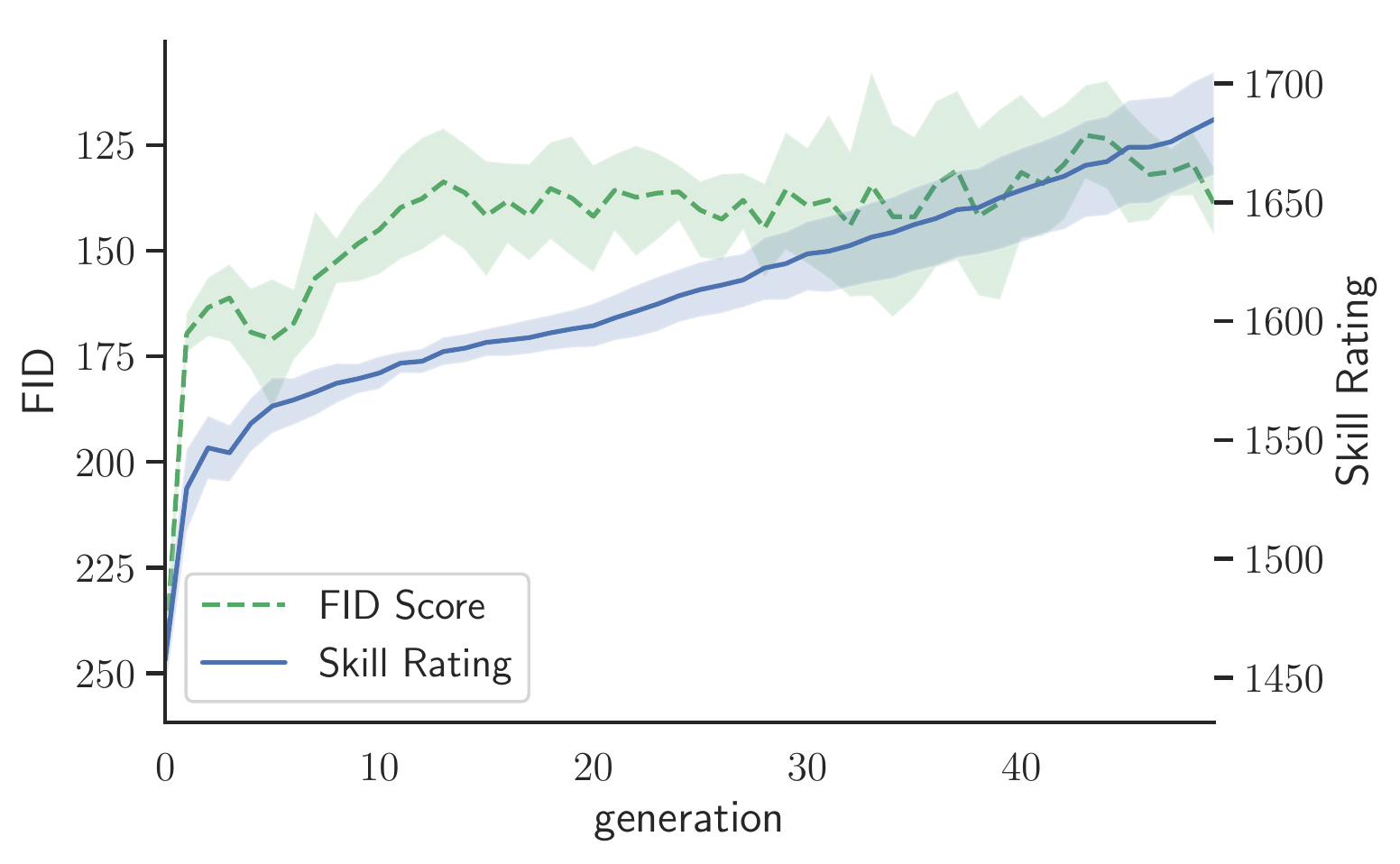}
		\caption{COEGAN + Skill, Pearson: -0.8, Spearman: -0.73}
		\label{fig:skill_skill_fid_comparison}
	\end{subfigure}\quad%
	\begin{subfigure}[t]{.45\textwidth}
		\centering
		\includegraphics[width=\textwidth]{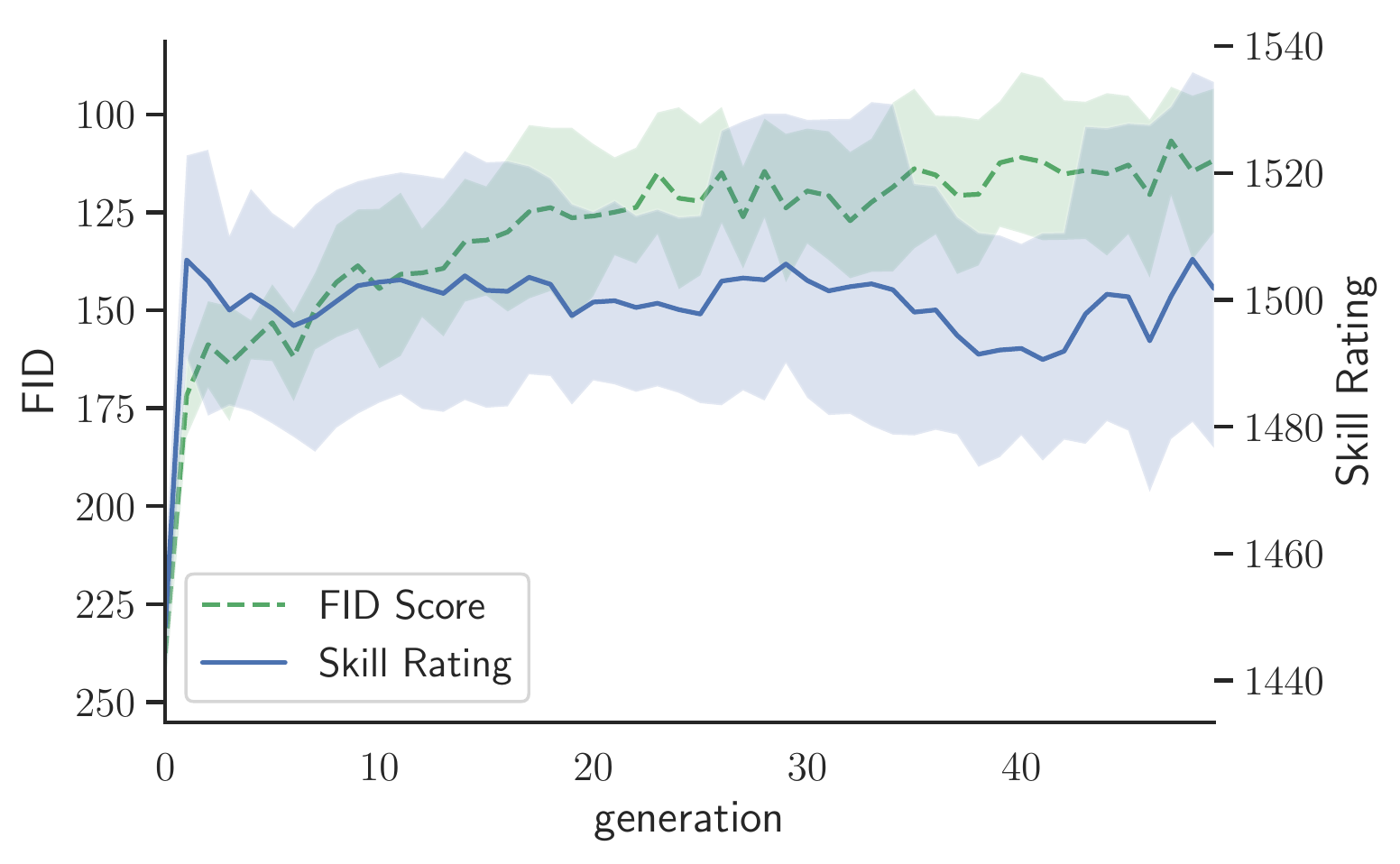}
		\caption{COEGAN + FID, Pearson: -0.54, Spearman: 0.18}
		\label{fig:fid_skill_fid_comparison}
	\end{subfigure}
	\begin{subfigure}[t]{.45\textwidth}
		\centering
		\includegraphics[width=\textwidth]{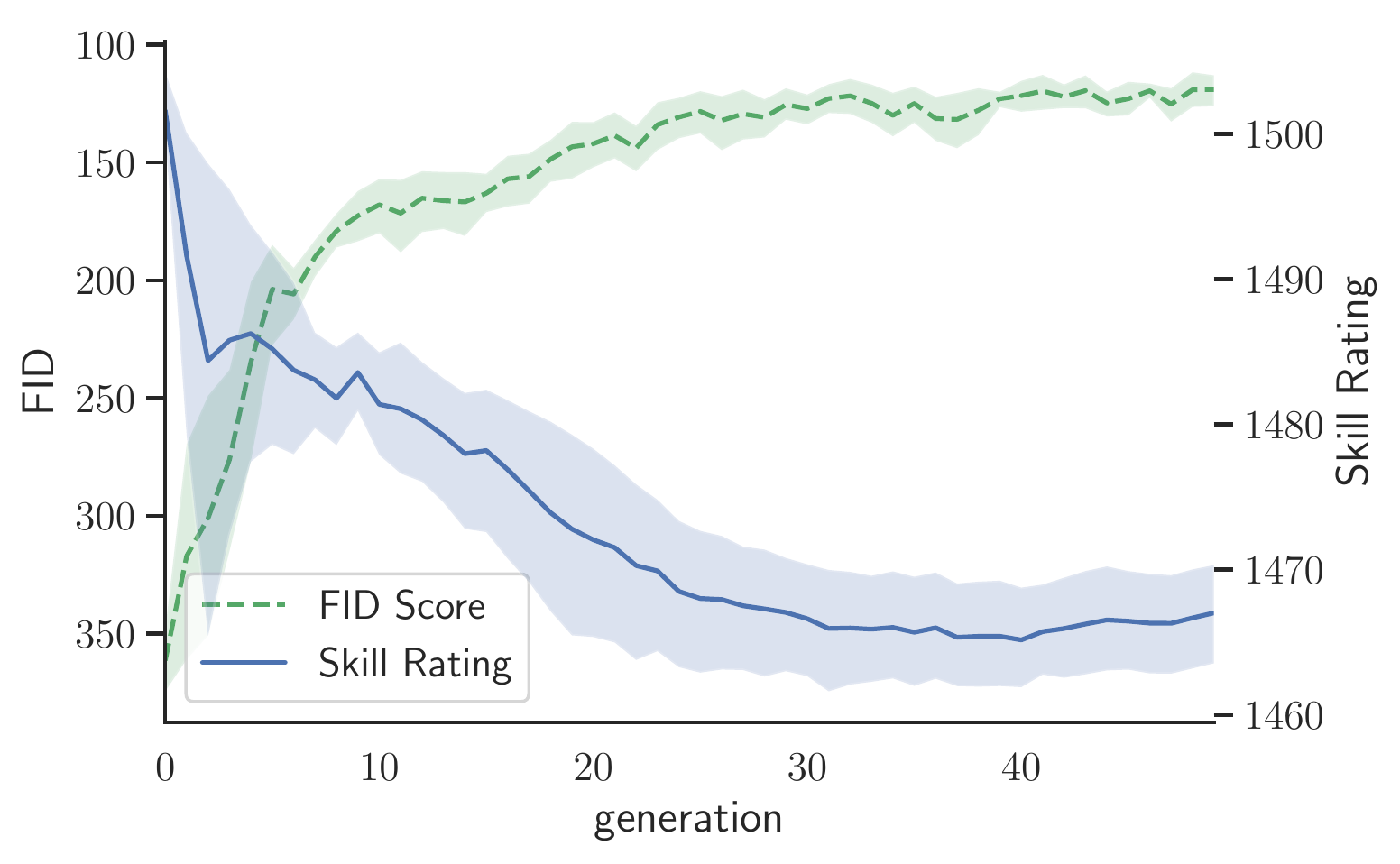}
		\caption{DCGAN-based, Pearson: 0.91, Spearman: 0.89}
		\label{fig:dcgan_skill_fid_comparison}
	\end{subfigure}\quad%
	\begin{subfigure}[t]{.45\textwidth}
		\centering
		\includegraphics[width=\textwidth]{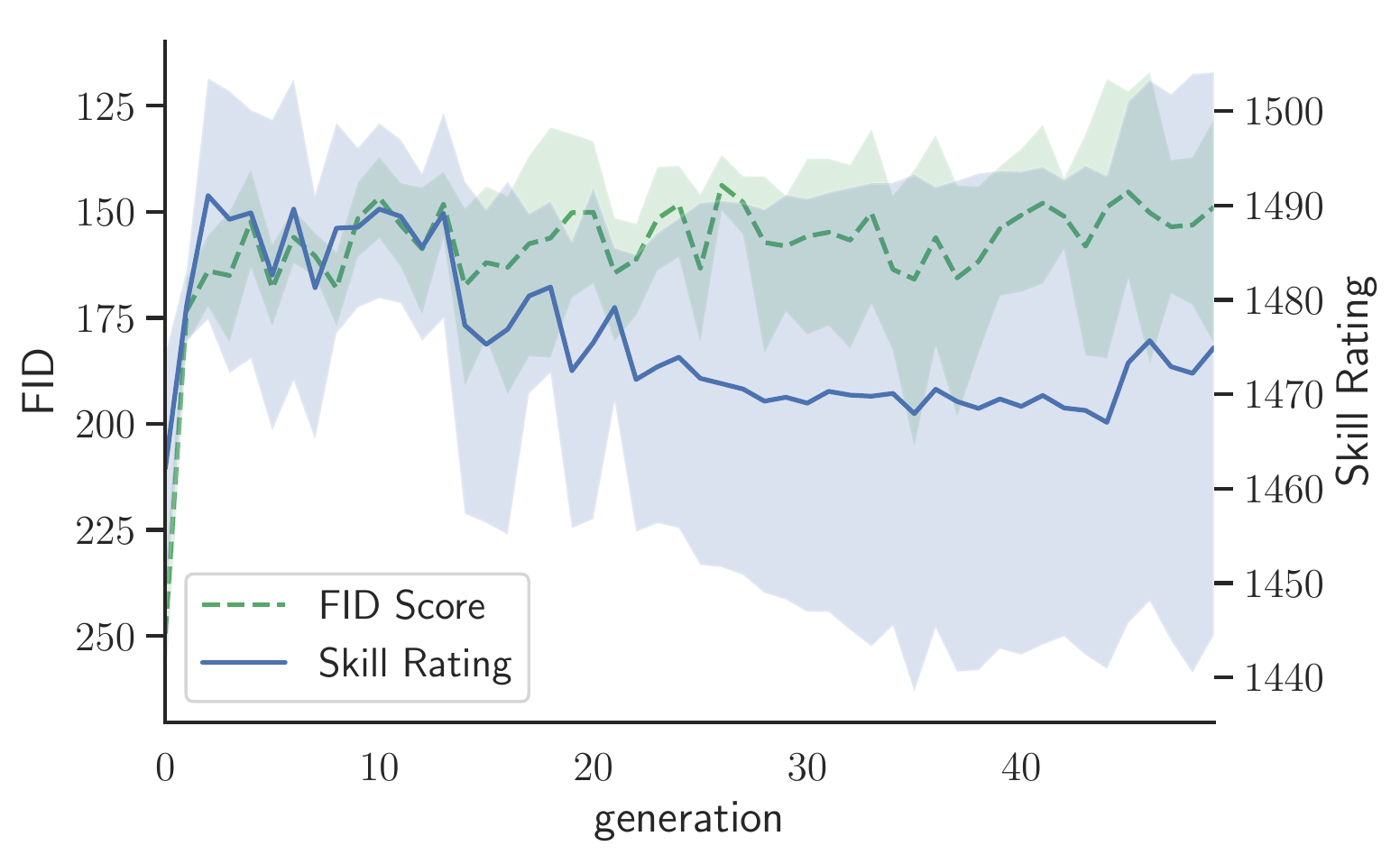}
		\caption{Random search, Pearson: -0.16, Spearman: 0.02}
		\label{fig:random_skill_fid_comparison}
	\end{subfigure}
	\caption{Comparison between the best FID score and the respective skill rating of generators trained with the SVHN dataset.}
	\label{fig:skill_fid_comparison}
\end{figure}

Figure~\ref{fig:skill_fid_comparison} shows the progression of the skill rating through generations compared with the best FID scores.
We can see in COEGAN guided by skill rating a clear improvement of the rating, as this is the same function used to provide evolutionary pressure in the individuals.
In the experiments of COEGAN with FID, the progress also exists but is less relevant.
The random approach presented an erratic behavior of the skill rating, showing that the individuals do not improve in this approach.
In the DCGAN-based experiments, the skill rating behaves differently, showing a decreasing pattern.
As there is only a single discriminator and generator, the number of matches per generation is only one.
Therefore, we do not meet the recommendations of the Glicko-2 system of having at least ten matches per time period and the rating is not useful for this case.

Except for the DCGAN experiments, we can also see in Figure~\ref{fig:skill_fid_comparison} some level of correlation between the best FID score and the respective skill rating among the generators in the populations.
The results demonstrated that skill rating follows the tendency of the FID score, evidencing that it can be used to guide the evolution of GANs.
We computed the Pearson correlation and the Spearman rank correlation between FID and skill rating to support this analysis.
We found a relevant negative correlation for the experiments with COEGAN guided by skill rating, achieving a Pearson correlation coefficient of $-0.8$ and a Spearman rank correlation of $-0.73$.
As FID is a distance measurement (lower is better) and skill rating is a score (high is better), the negative correlation is expected.

\begin{figure}[ht]
	\centering
	\begin{subfigure}[t]{.45\textwidth}
		\centering
		\includegraphics[width=\textwidth]{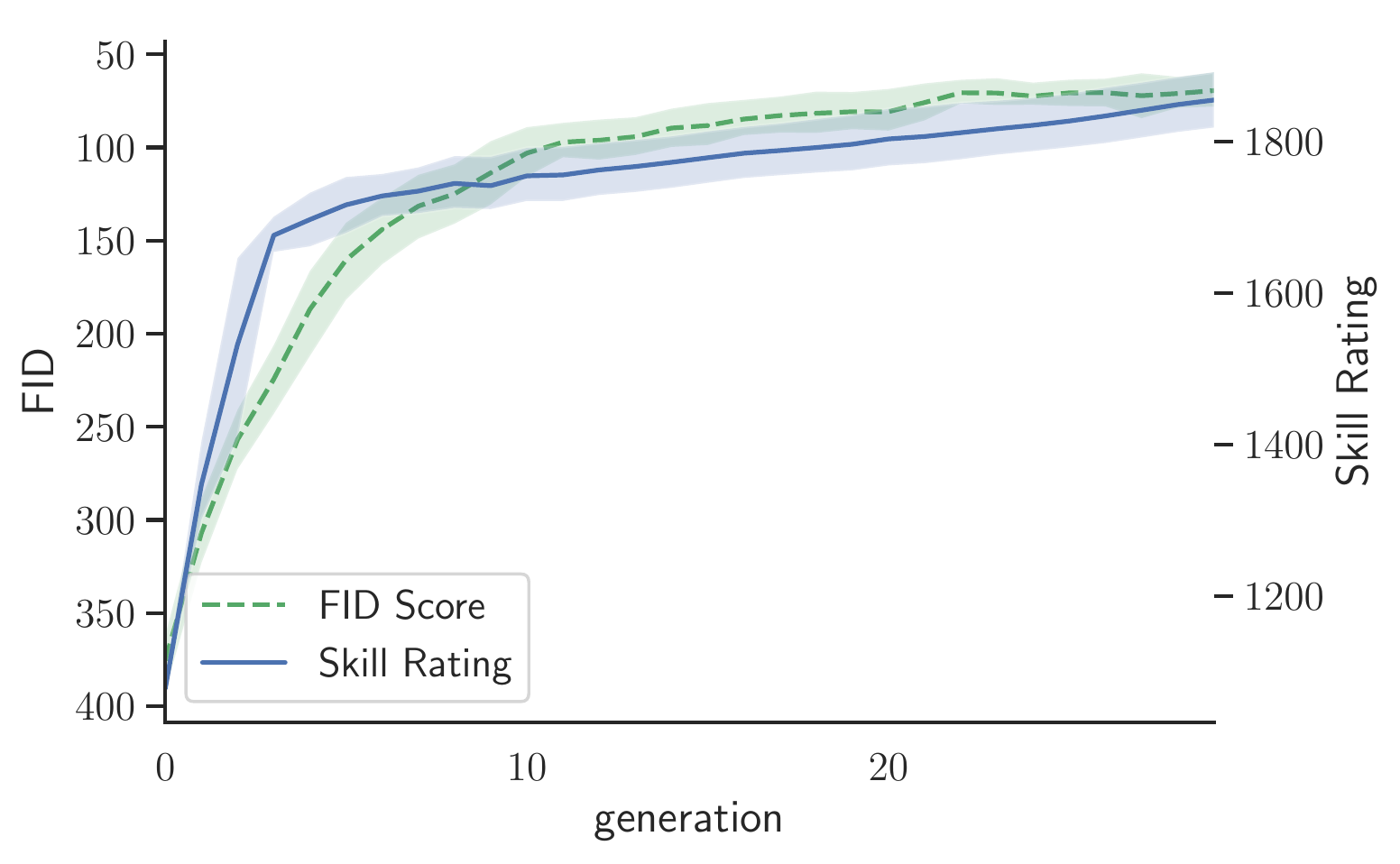}
		\caption{Best FID score and the respective skill rating for COEGAN + Skill. Pearson: $-0.96$, Spearman: $-0.99$}
		\label{fig:mnist_skill_fid_comparison}
	\end{subfigure}\qquad%
	\begin{subfigure}[t]{.45\textwidth}
		\centering
		\includegraphics[width=\textwidth]{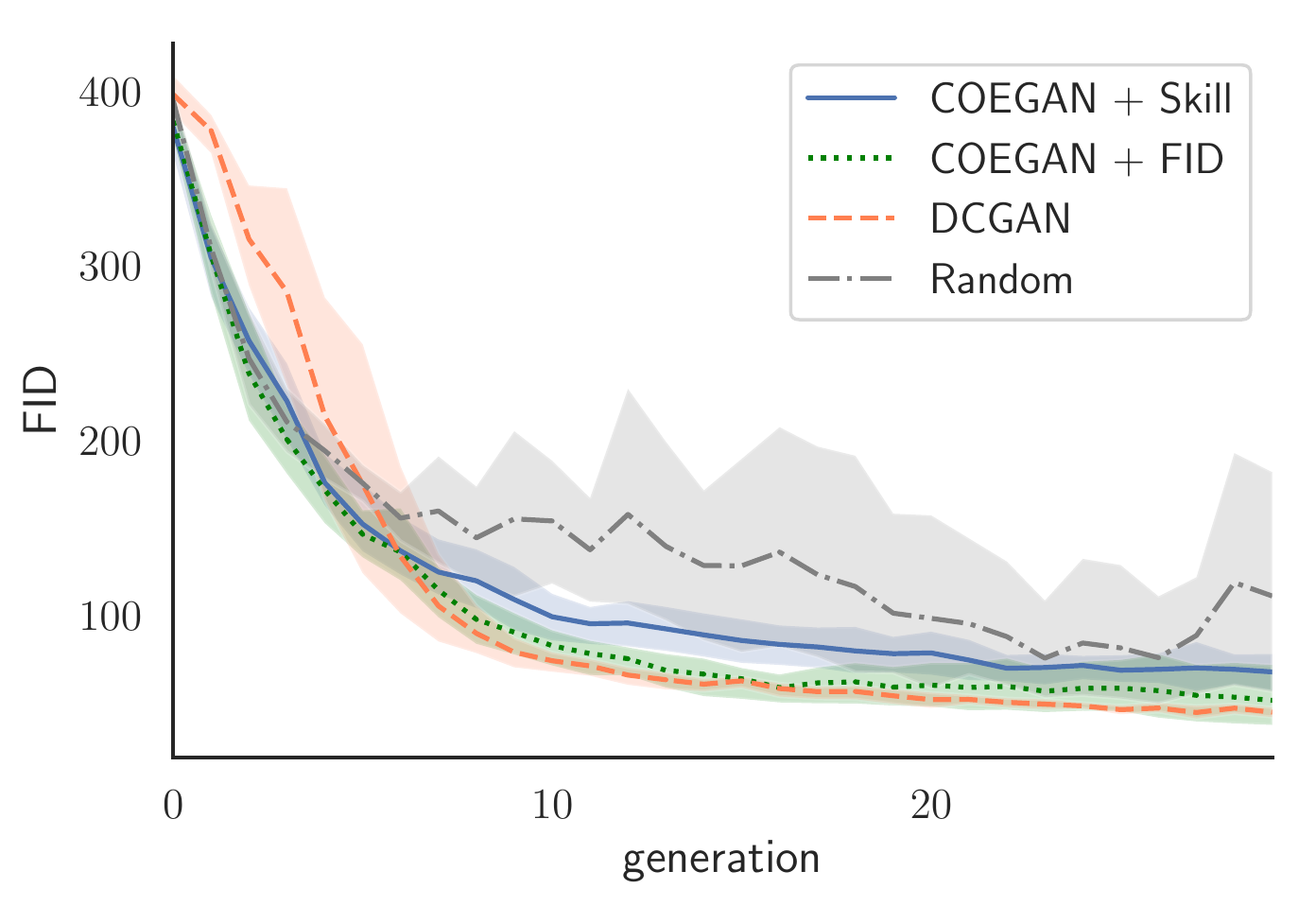}
		\caption{Best FID score for all solutions}
		\label{fig:mnist_fid}
	\end{subfigure}
	\caption{Results for the experiments with the MNIST dataset.}
	\label{fig:results_mnist}
\end{figure}

We experienced high variability on the FID score in the experiments with the SVHN dataset, both for the Inception Network trained with the ImageNet and SVHN datasets.
Therefore, we conduct a study using the MNIST dataset to enhance the relationship between the FID score and skill rating.
We followed the same parameters presented in Table~\ref{table:setup}, but limiting the number of generations to $30$.
Figure~\ref{fig:mnist_skill_fid_comparison} shows a smoother progression of skill rating and FID, illustrating a more clear relation between them, which is evidenced by the Pearson's correlation coefficient of $-0.96$ and the Spearman's rank correlation of $-0.99$.
We also show in Figure~\ref{fig:mnist_fid} that COEGAN guided by skill rating achieves performance similar to COEGAN guided by FID, outperforming the random search approach.

\begin{figure}[ht]
	\centering
	\begin{subfigure}[t]{.5\textwidth}
		\centering
		\includegraphics[width=\textwidth]{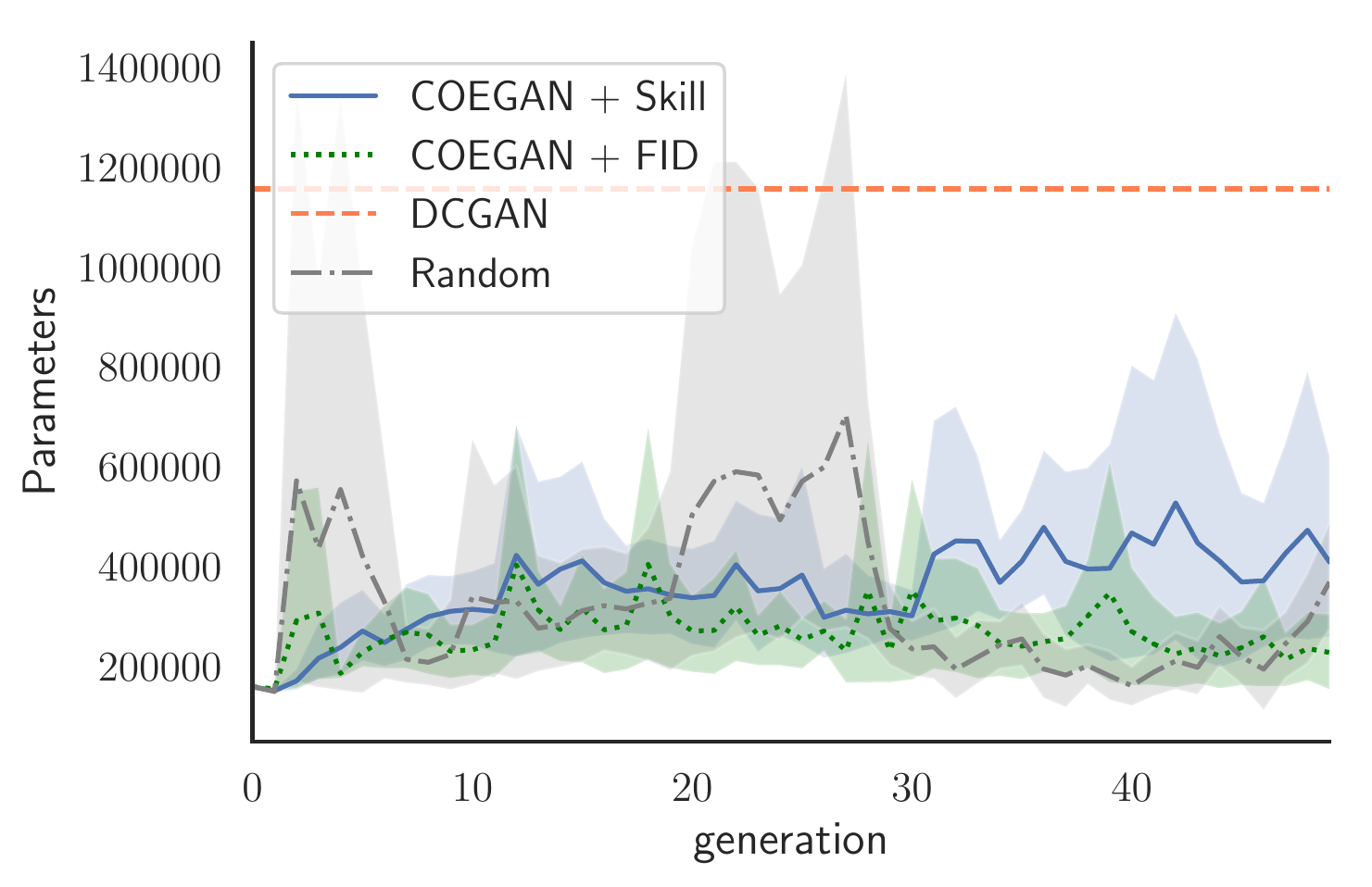}
		\caption{Number of parameters for generators}
		\label{fig:parameters_g}
	\end{subfigure}%
	\begin{subfigure}[t]{.5\textwidth}
		\centering
		\includegraphics[width=\textwidth]{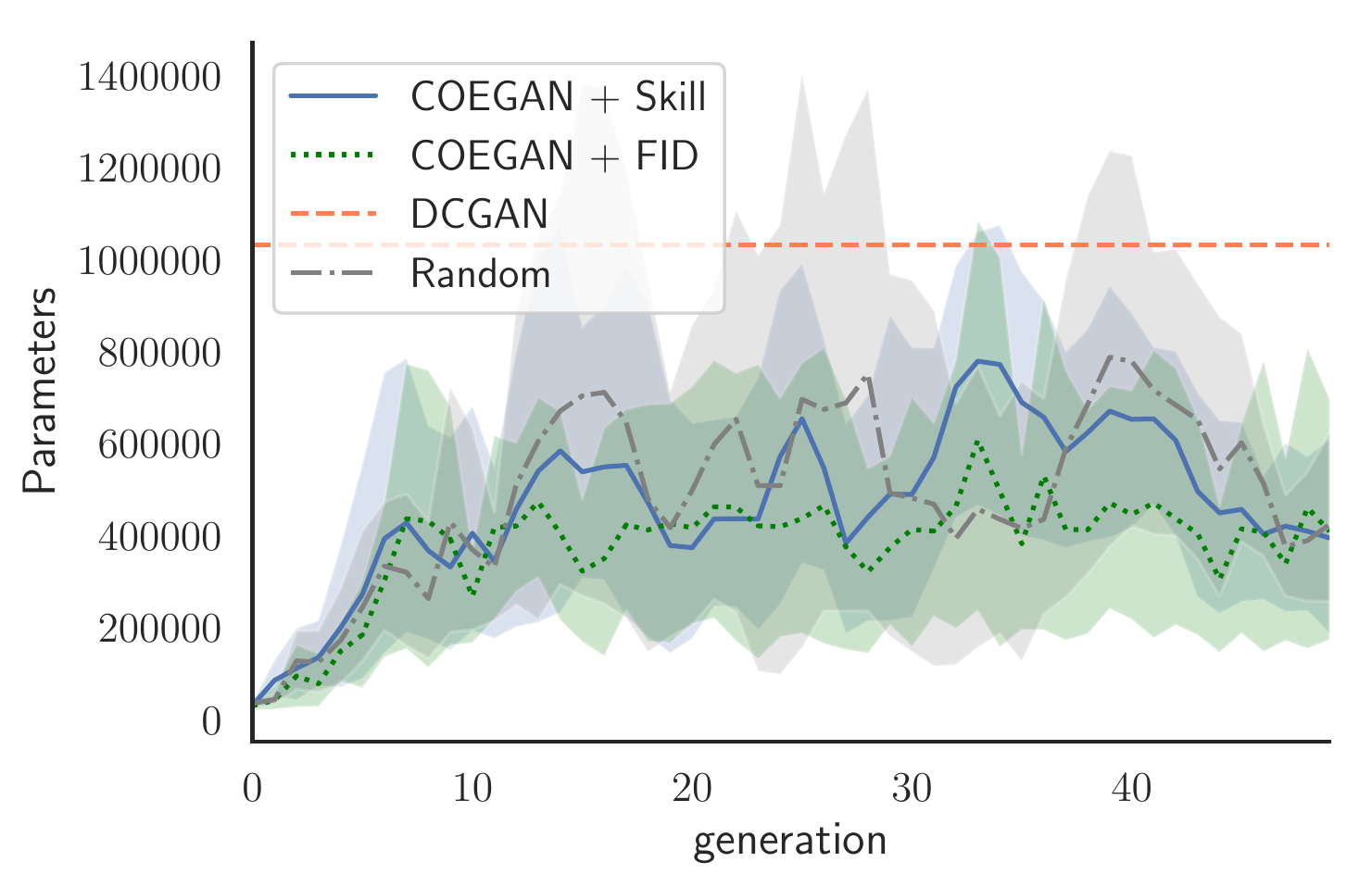}
		\caption{Number of parameters for discriminators}
		\label{fig:parameters_d}
	\end{subfigure}
	\caption{Average number of parameters in the neural networks of generators and discriminators at each generation. Note that the number of parameters for the DCGAN-based experiments is constant, as there is not an evolutionary algorithm applied to this case.}
	\label{fig:parameters}
\end{figure}

Figure~\ref{fig:parameters} presents the average number of parameters in generators and discriminators from the experiments with the SVHN dataset.
As there is no evolutionary algorithm applied to DCGAN, the number of parameters is constant.
It is important to note that the average number of parameters on the individuals in the COEGAN experiments is much lower than the parameters in DCGAN.
Despite this, the results of COEGAN are still better than DCGAN.
Therefore, the experiments evidenced that the evolutionary algorithm applied in COEGAN was able to find more efficient models.
We limited in the experimental setup the complexity and the number of layers in the genome.
Experiments with an expanded setup should be conducted to assess the possibility of even better results.

\begin{figure}[ht]
	\centering
	\begin{subfigure}[t]{.45\textwidth}
		\centering
		\includegraphics[width=\textwidth]{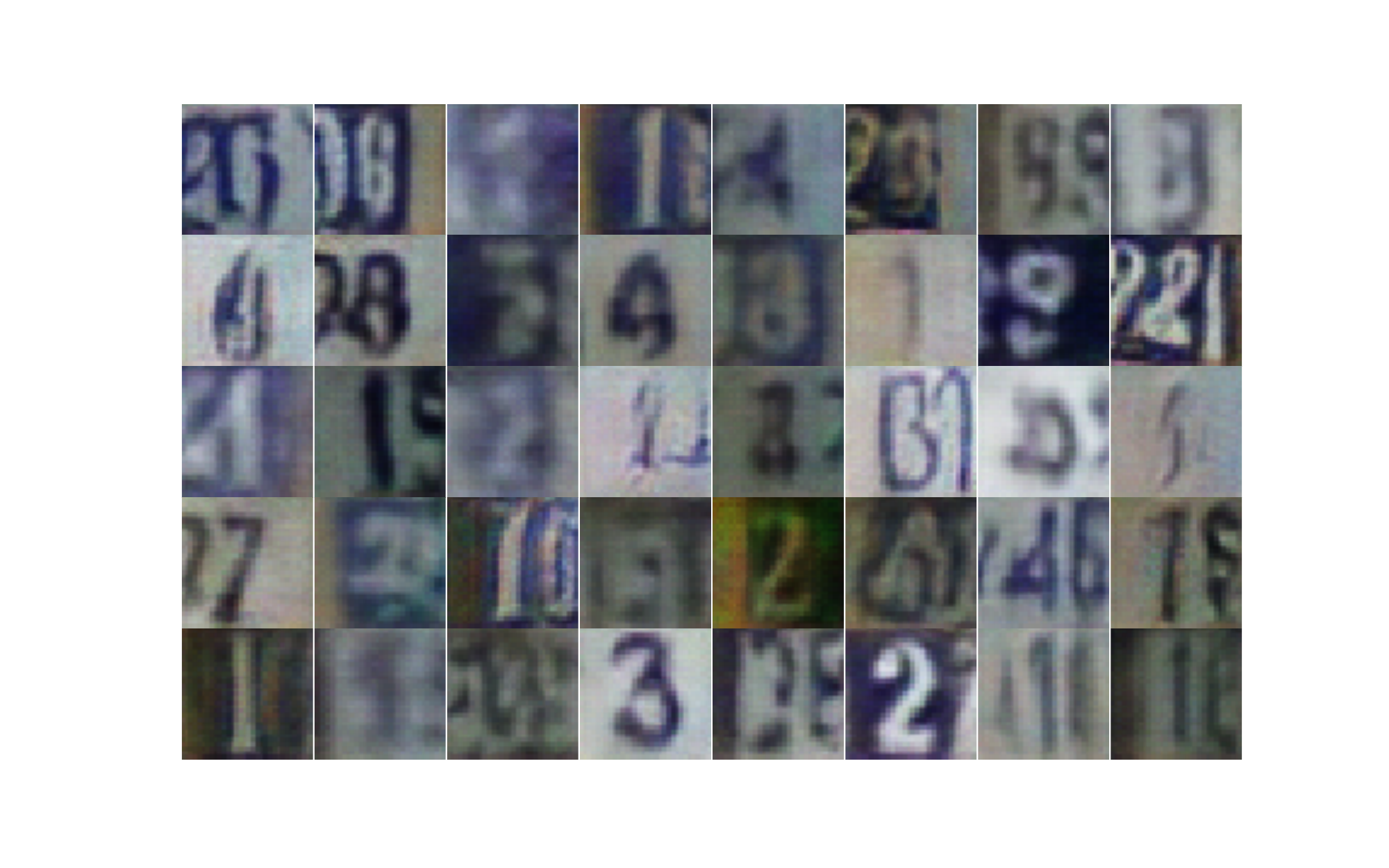}
		\caption{COEGAN with skill rating as fitness}
		\label{fig:svhn_samples}
	\end{subfigure}\qquad%
	\begin{subfigure}[t]{.45\textwidth}
		\centering
		\includegraphics[width=\textwidth]{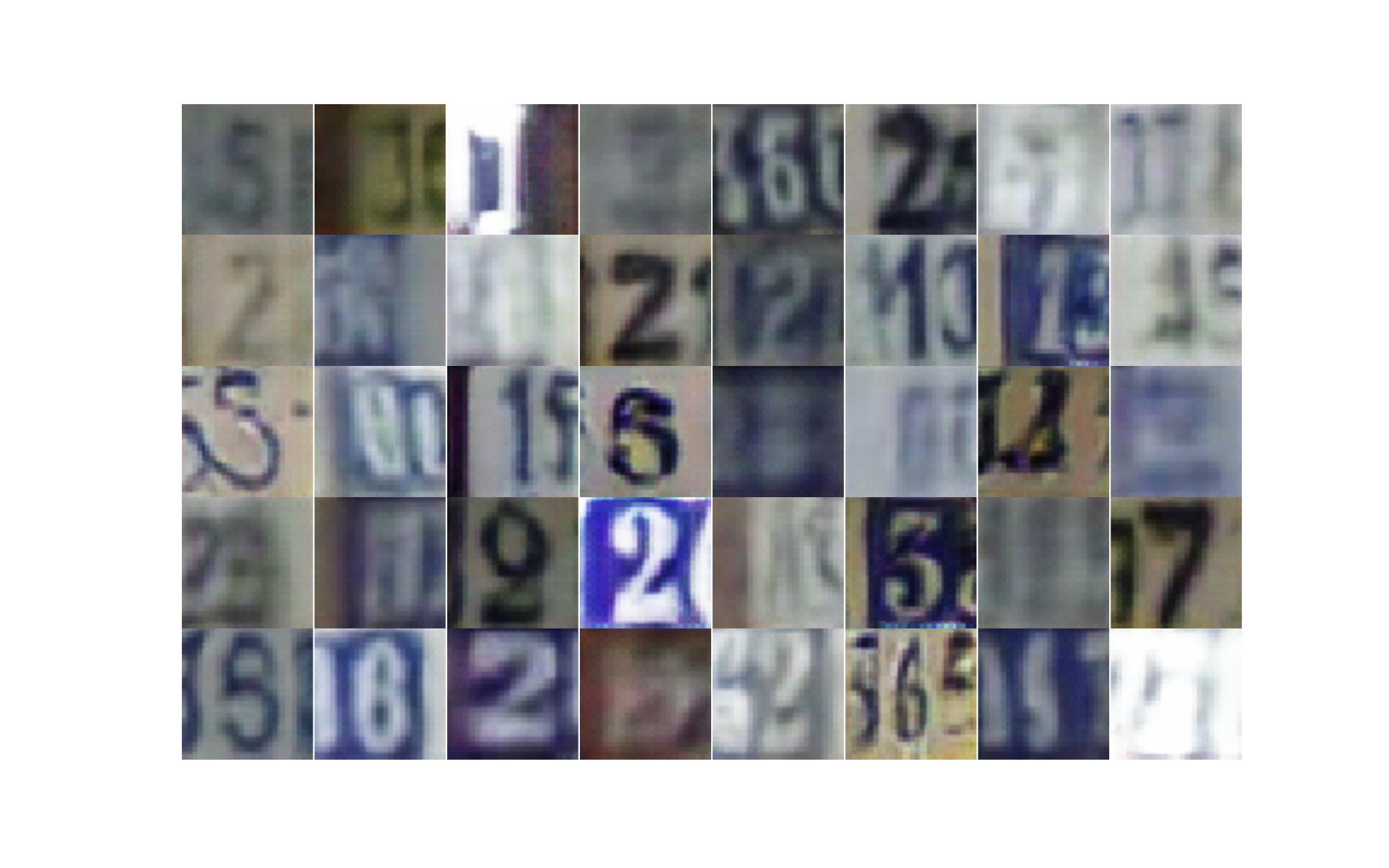}
		\caption{COEGAN with the FID score and loss function as fitness}
		\label{fig:svhn_samples}
	\end{subfigure}
	\caption{Samples produced by the best generator after the COEGAN training.}
	\label{fig:svhn_samples}
\end{figure}

Figure~\ref{fig:svhn_samples} shows samples produced by the generator after the COEGAN training with FID and skill rating as fitness.
In order to achieve better quality, we trained the algorithms using $200$ batches at each generation (instead of $20$).
We can see that the quality of the samples is similar, with both strategies presenting variability on the samples.

\section{Conclusions}
\label{sec:conclusions}

Generative Adversarial Networks (GANs) represented a relevant advance in generative models, producing impressive results in contexts such as the image domain.
In spite of this, the training of a GAN is challenging and often requires a trial-and-error approach to achieve the desired outcome.
Several strategies were used in order to improve training stability and produce better results.
Proposals modified the original GAN model to introduce alternative loss functions and architectural changes.
On the other hand, the use of evolutionary algorithms in the context of GANs was recently proposed.
COEGAN combines neuroevolution and coevolution on the training and evolution of GANs.
However, experiments identified that the fitness used in COEGAN can be improved to better guide the evolution of discriminators and generators in the populations.

We propose the use of a game rating system, based on the application of Glicko-2 introduced in~\cite{olsson2018skill}, to design a new fitness strategy for COEGAN.
Thus, we changed the fitness functions used by discriminators and generators to use the skill rating metric instead of the loss function and the FID score.
We conducted experiments to evaluate this proposal and compare the results with the previous COEGAN fitness proposal, a DCGAN-based approach, and a random search model.

The results evidenced that, although the FID score as fitness provides better results, the skill rating method also contribute with useful information in the evolution of GANs.
The use of COEGAN with skill rating outperforms the random search approach, demonstrating the effectiveness of this fitness function.
When compared to the FID score, the advantages when using skill rating is the lower execution cost and the self-contained solution, i.e., skill rating does not need to use an external component such as in the FID score.
The calculation of the FID requires a trained Inception Network, making the score highly dependent on the context where it was trained and applied.
Therefore, skill rating has the potential to be used in more domains.
Besides, the skill rating does not require a neural network to interpret images produced by generators.
Instead, the output of the discriminator is used in the calculation, resulting in a lower execution cost when compared to the FID score.
We also show that there is a correlation between the FID score and the skill rating metric when using the latter as fitness with COEGAN.
However, skill rating worked better with the MNIST dataset, making this correlation more evident.
The SVHN dataset is more complex and sometimes lead to disagreement between FID and skill rating.
The strategy to obtain the results of matches between generators and discriminators can be improved to better represent the player's skill.

As future work, we aim to expand the strategies evaluated in this paper regarding the use of skill rating as fitness.
We will evaluate changes in the skill tournament to take into account individuals from previous generations.
Besides, different strategies to calculate the outcome of matches can be used to improve the results.
We will investigate the use of strategies that bring information about the variability of the samples produced by generators, in order to approximate the information provided by the FID score.

\section*{Acknowledgments}\label{sec:acknowledgments}
This article is based upon work from COST Action CA15140: ImAppNIO, supported by COST (European Cooperation in Science and Technology).

\bibliographystyle{splncs04}
\bibliography{costa}

\begin{thebibliography}{10}
\providecommand{\url}[1]{\texttt{#1}}
\providecommand{\urlprefix}{URL }
\providecommand{\doi}[1]{https://doi.org/#1}

\bibitem{al2018towards}
Al-Dujaili, A., Schmiedlechner, T., Hemberg, E., O’Reilly, U.M.: Towards
  distributed coevolutionary {GANs}. In: AAAI 2018 Fall Symposium (2018)

\bibitem{antonio2018coevolutionary}
Antonio, L.M., Coello, C.A.C.: Coevolutionary multiobjective evolutionary
  algorithms: Survey of the state-of-the-art. IEEE Transactions on Evolutionary
  Computation  \textbf{22}(6),  851--865 (2018)

\bibitem{arjovsky2017wasserstein}
Arjovsky, M., Chintala, S., Bottou, L.: Wasserstein generative adversarial
  networks. In: International Conference on Machine Learning. pp. 214--223
  (2017)

\bibitem{borji2019pros}
Borji, A.: Pros and cons of {GAN} evaluation measures. Computer Vision and
  Image Understanding  \textbf{179},  41--65 (2019)

\bibitem{costa2019evaluating}
Costa, V., Louren{\c{c}}o, N., Correia, J., Machado, P.: {COEGAN}: Evaluating
  the coevolution effect in generative adversarial networks. In: Proceedings of
  the Genetic and Evolutionary Computation Conference. pp. 374--382. ACM (2019)

\bibitem{costa2019coevolution}
Costa, V., Louren{\c{c}}o, N., Machado, P.: Coevolution of generative
  adversarial networks. In: International Conference on the Applications of
  Evolutionary Computation (Part of EvoStar). pp. 473--487. Springer (2019)

\bibitem{garciarena2018evolved}
Garciarena, U., Santana, R., Mendiburu, A.: Evolved {GANs} for generating
  pareto set approximations. In: Proceedings of the Genetic and Evolutionary
  Computation Conference. pp. 434--441. GECCO '18, ACM, New York, NY, USA
  (2018)

\bibitem{glickman2012example}
Glickman, M.E.: Example of the glicko-2 system. Boston University pp.~1--6
  (2013), \url{http://www.glicko.net/glicko/glicko2.pdf}

\bibitem{NIPS2014_5423}
Goodfellow, I., Pouget-Abadie, J., Mirza, M., Xu, B., Warde-Farley, D., Ozair,
  S., Courville, A., Bengio, Y.: Generative adversarial nets. In: NIPS. Curran
  Associates, Inc. (2014)

\bibitem{gulrajani2017improved}
Gulrajani, I., Ahmed, F., Arjovsky, M., Dumoulin, V., Courville, A.C.: Improved
  training of wasserstein {GANs}. In: Advances in Neural Information Processing
  Systems. pp. 5769--5779 (2017)

\bibitem{heusel2017gans}
Heusel, M., Ramsauer, H., Unterthiner, T., Nessler, B., Hochreiter, S.: {GANs}
  trained by a two time-scale update rule converge to a local nash equilibrium.
  In: Advances in Neural Information Processing Systems. pp. 6629--6640 (2017)

\bibitem{jolicoeur-martineau2018}
Jolicoeur-Martineau, A.: The relativistic discriminator: a key element missing
  from standard {GAN}. In: International Conference on Learning Representations
  (2019)

\bibitem{karras2018progressive}
Karras, T., Aila, T., Laine, S., Lehtinen, J.: Progressive growing of {GAN}s
  for improved quality, stability, and variation. In: International Conference
  on Learning Representations (2018)

\bibitem{kingma2015adam}
Kingma, D.P., Ba, J.: Adam: A method for stochastic optimization. In:
  International Conference on Learning Representations (ICLR) (2015)

\bibitem{lecun1998mnist}
LeCun, Y.: The mnist database of handwritten digits. http://yann. lecun.
  com/exdb/mnist/  (1998)

\bibitem{mao2017least}
Mao, X., Li, Q., Xie, H., Lau, R.Y., Wang, Z., Smolley, S.P.: Least squares
  generative adversarial networks. In: 2017 IEEE International Conference on
  Computer Vision (ICCV). pp. 2813--2821. IEEE (2017)

\bibitem{miikkulainen2017evolving}
Miikkulainen, R., Liang, J., Meyerson, E., Rawal, A., Fink, D., Francon, O.,
  Raju, B., Navruzyan, A., Duffy, N., Hodjat, B.: Evolving deep neural
  networks. arXiv preprint arXiv:1703.00548  (2017)

\bibitem{mitchell2006coevolutionary}
Mitchell, M.: Coevolutionary learning with spatially distributed populations.
  Computational intelligence: principles and practice  (2006)

\bibitem{netzer2011reading}
Netzer, Y., Wang, T., Coates, A., Bissacco, A., Wu, B., Ng, A.Y.: Reading
  digits in natural images with unsupervised feature learning  (2011)

\bibitem{olsson2018skill}
Olsson, C., Bhupatiraju, S., Brown, T., Odena, A., Goodfellow, I.: Skill rating
  for generative models. arXiv preprint arXiv:1808.04888  (2018)

\bibitem{radford2015unsupervised}
Radford, A., Metz, L., Chintala, S.: Unsupervised representation learning with
  deep convolutional generative adversarial networks. arXiv preprint
  arXiv:1511.06434  (2015)

\bibitem{russakovsky2015imagenet}
Russakovsky, O., Deng, J., Su, H., Krause, J., Satheesh, S., Ma, S., Huang, Z.,
  Karpathy, A., Khosla, A., Bernstein, M., et~al.: Imagenet large scale visual
  recognition challenge. International Journal of Computer Vision
  \textbf{115}(3),  211--252 (2015)

\bibitem{salimans2016improved}
Salimans, T., Goodfellow, I., Zaremba, W., Cheung, V., Radford, A., Chen, X.:
  Improved techniques for training {GANs}. In: Advances in Neural Information
  Processing Systems. pp. 2234--2242 (2016)

\bibitem{sims1994evolving}
Sims, K.: Evolving 3d morphology and behavior by competition. Artificial life
  \textbf{1}(4),  353--372 (1994)

\bibitem{neat}
Stanley, K.O., Miikkulainen, R.: Evolving neural networks through augmenting
  topologies. Evolutionary computation  \textbf{10}(2),  99--127 (2002)

\bibitem{stanley2004competitive}
Stanley, K.O., Miikkulainen, R.: Competitive coevolution through evolutionary
  complexification. Journal of Artificial Intelligence Research  \textbf{21},
  63--100 (2004)

\bibitem{szegedy2015going}
Szegedy, C., Liu, W., Jia, Y., Sermanet, P., Reed, S., Anguelov, D., Erhan, D.,
  Vanhoucke, V., Rabinovich, A.: Going deeper with convolutions. In:
  Proceedings of the IEEE conference on computer vision and pattern
  recognition. pp.~1--9 (2015)

\bibitem{szegedy2016rethinking}
Szegedy, C., Vanhoucke, V., Ioffe, S., Shlens, J., Wojna, Z.: Rethinking the
  inception architecture for computer vision. In: Proceedings of the IEEE
  Conference on Computer Vision and Pattern Recognition. pp. 2818--2826 (2016)

\bibitem{toutouh2019spatial}
Toutouh, J., Hemberg, E., O’Reilly, U.M.: Spatial evolutionary generative
  adversarial networks. arXiv preprint arXiv:1905.12702  (2019)

\bibitem{vevcek2014comparison}
Ve{\v{c}}ek, N., {\v{C}}repin{\v{s}}ek, M., Mernik, M., Hrn{\v{c}}i{\v{c}}, D.:
  A comparison between different chess rating systems for ranking evolutionary
  algorithms. In: 2014 Federated Conference on Computer Science and Information
  Systems. pp. 511--518. IEEE (2014)

\bibitem{vevcek2014chess}
Ve{\v{c}}ek, N., Mernik, M., {\v{C}}repin{\v{s}}ek, M.: A chess rating system
  for evolutionary algorithms: A new method for the comparison and ranking of
  evolutionary algorithms. Information Sciences  \textbf{277},  656--679 (2014)

\bibitem{wang2018evolutionary}
Wang, C., Xu, C., Yao, X., Tao, D.: Evolutionary generative adversarial
  networks. arXiv preprint arXiv:1803.00657  (2018)

\bibitem{xu2018empirical}
Xu, Q., Huang, G., Yuan, Y., Guo, C., Sun, Y., Wu, F., Weinberger, K.: An
  empirical study on evaluation metrics of generative adversarial networks.
  arXiv preprint arXiv:1806.07755  (2018)

\bibitem{zhang2018self}
Zhang, H., Goodfellow, I., Metaxas, D., Odena, A.: Self-attention generative
  adversarial networks. arXiv preprint arXiv:1805.08318  (2018)

\end{thebibliography}

\end{document}